\documentclass[journal]{IEEEtran}
\usepackage{amsmath,amsfonts}
\usepackage{algorithmic}
\usepackage{algorithm}
\pdfoutput=1
\usepackage[utf8]{inputenc}  
\usepackage{amsmath}  
\usepackage{graphicx} 
\usepackage{CJKutf8}  
\usepackage{fontenc}  
\usepackage{array}
\usepackage[caption=false,font=normalsize,labelfont=sf,textfont=sf]{subfig}
\usepackage{amsmath} 
\usepackage{textcomp}
\usepackage{stfloats}
\usepackage[utf8]{inputenc}
\usepackage{url}
\usepackage{graphicx}
\usepackage[T1]{fontenc}
\usepackage{float}
\usepackage[table,xcdraw]{xcolor}
\usepackage{colortbl} 
\usepackage{verbatim}
\usepackage{multirow}
\usepackage{graphicx}
\usepackage{longtable}
\usepackage{cite}

\usepackage[colorlinks,
            linkcolor=blue,
            anchorcolor=blue,
            citecolor=blue]{hyperref}
\begin{document}

\title{Text-guided Zero-Shot Object Localization }

\author{Jingjing Wang, Xinglin Piao, Zongzhi Gao, Bo Li, Yong Zhang$^{*}$, Baocai Yin}
        % <-this % stops a space

\markboth{Journal of \LaTeX\ Class Files,~Vol.~xx, No.~x, Nov.2024}%
{Shell \MakeLowercase{\textit{et al.}}: A Sample Article Using IEEEtran.cls for IEEE Journals}


\maketitle

\begin{abstract}
Object localization is a hot issue in computer vision area, which aims to identify and determine the precise location of specific objects from image or video. Most existing object localization methods heavily rely on extensive labeled data, which are costly to annotate and constrain their applicability. Therefore, we propose a new Zero-Shot Object Localization (ZSOL) framework for addressing the aforementioned challenges. In the proposed framework, we introduce the Contrastive Language-Image Pre-training (CLIP) module which could integrate visual and linguistic information effectively. Furthermore, we design a Text Self-Similarity Matching (TSSM) module, which could improve the localization accuracy by enhancing the representation of text features extracted by CLIP module. Hence, the proposed framework can be guided by prompt words to identify and locate specific objects in an image in the absence of labeled samples. The results of extensive experiments demonstrate that the proposed method could improve the localization performance significantly and establishes an effective benchmark for further research.

% \Piao{Object localization is a hot issue in computer vision area, which aims to identify and determine the precise location of specific objects from image or video. Most existing object localization methods heavily rely on extensive labeled data. However, the annotation of these data is resource-intensive and constrains the overall applicability of such methods. Therefore, we propose a new Zero-Shot Object Localization (ZSOL) framework for addressing the aforementioned challenges. In the proposed framework, we introduce the Contrastive Language-Image Pre-training (CLIP) module which could integrate visual and linguistic information effectively. Furthermore, we design a Text Self-Similarity Matching (TSSM) module which could improve the localization accuracy by enhancing the representation of text features extracted by CLIP module. Hence, the proposed framework can be guided by prompt words to identify and locate specific objects in an image in the absence of labeled samples. The results of extensive experiments demonstrate that the proposed method could improve the localization performance significantly and establishes an effective benchmark for further research.} 

\end{abstract}


\section{Introduction}
\IEEEPARstart{O}{bject} localization is one of the fundamental and challenging problems in computer vision and has attracted considerable attention from various industries. The goal of object localization is to accurately determine the location of specific objects in an image or video. Object localization is crucial in many applications, including the fields of object detection \cite{1,2,3}, semantic segmentation \cite{4,5}, and object tracking \cite{6,7,csvt1}. The accuracy of object localization is directly related to the effectiveness and practicality of the system in many applications such as autonomous driving \cite{8,9}, video surveillance\cite{10,11}, and medical image analysis \cite{12}. 

% \sout{Therefore, in-depth research and continuous optimization of object localization technology is of great importance to improve the level of automation and accelerate the process of interdisciplinary research.}

\begin{figure}[!t]
\centering
\includegraphics[width=0.5\textwidth]{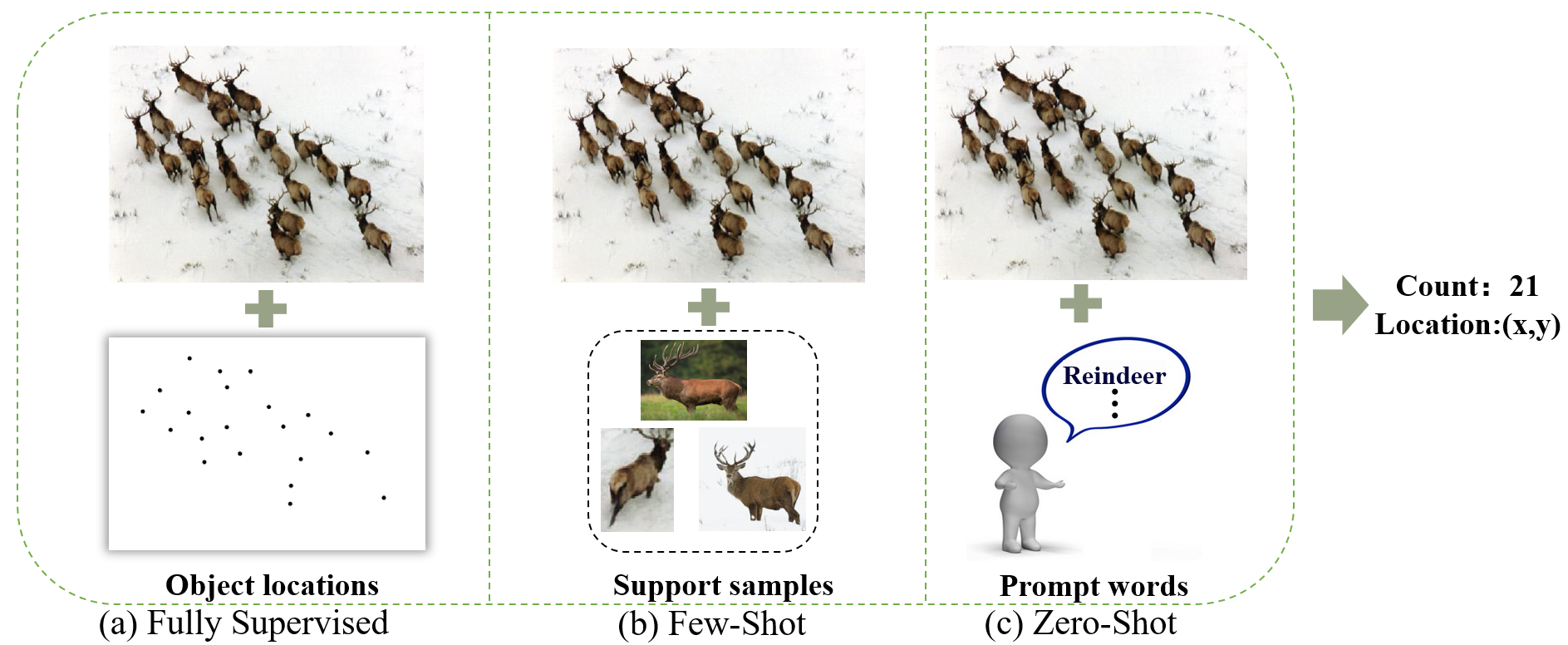}
\caption{Paradigm of object localization. (a) Fully Supervised: Costly due to the need for extensive manual labeling. (b) Few-Shot: Highly dependent on the quality of the support samples. (c) Zero-Shot: No annotation, text prompts guide model image understanding.}
\label{fig_1}
\end{figure}

In past decades, a series of methods have been proposed for object localization. These methods generally rely on a large number of labeled data to assist the model in learning features, as shown in Fig. \ref{fig_1}-(a). However, obtaining large-scale labeled object localization data is often a time-consuming and laborious task. Therefore, Few-Shot Object Localization (FSOL) task have emerged to provide researchers with the opportunity to perform effective model training and evaluation in resource-constrained environments \cite{13,14,15}. This approach is not only cost-effective, but also better adapted to the needs of specific domains. However, due to the scarcity of training samples, the dependence of few-shot localization on data is greatly increased, as shown in Fig. \ref{fig_1}-(b). Few-shot based object localization methods usually need to learn the features and distributions of the object categories from a small amount of labeled data, which may have the risk of overfitting. Especially when the quality of labeled data is not high or the category samples are not balanced, few-shot based object localization methods may fail to perform effectively. In addition, few-shot based object localization methods require manually annotated high-quality samples \cite{Kang} and may have difficulty with similar classes or noisy data \cite{17}. Compared to few-shot learning, zero-shot learning could potentially offer enhanced flexibility and versatility. This is because zero-shot learning allows models to generalize beyond the specific examples they have been trained on, enabling them to recognize and understand new categories or concepts that were not part of their initial training data. Therefore, Zero-Shot Object Localization (ZSOL) is more suitable for the situation of labeled data scarcity.
% \sout{Therefore, careful consideration needs to be given to data quality, category characteristics, and the generalization ability of the model when using few-shot localization methods.} 
% \sout{The difference is that zero-shot object localization does not require the labeling information of the object samples and possesses excellent cross-domain migration capability,}
Zero-shot based object localization method does not require the labeling information of the object samples and could possess excellent cross-domain migration as shown in Fig. \ref{fig_1}-(c). However, zero-shot object localization also faces some problems such as inter-category semantic differences and visual variations.

For the above problems, Contrastive Language Image Pre-training (CLIP) model \cite{18} is a promising approach. The model is capable of mapping images and text into a shared vector space, thus enabling unsupervised joint learning between images and text. It is able to understand the semantic relationships between images and text. Meanwhile, CLIP could also generalize to unseen categories and scenarios effectively. However, it is important to note that textual descriptions have certain limitations in conveying image related information. Compared to the direct elaboration of visual features such as shape and color, textual descriptions focus more on the expression of abstract and implicit characteristics, which may lead to a certain deviation from the actual characteristics of the object. More critically, images contain much more information than textual descriptions can cover, a difference that can lead to ambiguity and confusion, and thus pose a major challenge in locating the object and all its details accurately in the image.

In this work, we propose a novel zero-shot object localization framework that aims to address the challenges inherent in object localization task through a cross-modal approach. The core concept of our model is to use linguistic text as flexible prompt words, which are converted into query vectors that are aligned with image features. This approach effectively mitigates the potential impact of semantic gaps and visual variations on localization accuracy. This enables our approach to accurately localize objects localization of interest in images without category-specific training data. Meanwhile, two complementary avenues are pursued with a view to offsetting the detrimental impact of semantic bias. On the one hand, by fine-tuning the prompt words from broad category descriptions (e.g., “fruits”) to more specific example descriptions (e.g., “grapes”), the model is able to capture the features of the object more accurately. On the other hand, inspired by the SSP \cite{ssp} approach, a text self-similarity matching module (TSSM) is designed to enhance the representational capacity of text features. This module further refines and enriches the features, thereby obtaining more comprehensive information for object localization.

In conclusion, our main contributions are as follows:

\begin{itemize}
    \item A novel cross-modal object localization framework is proposed that achieves zero-shot localization using prompt words for the first time.

    \item The designed TSSM module improves localization accuracy by enhancing the representational capacity  for textual features.

    \item Extensive experiments on multiple datasets show that our ZSOL approach achieves high performance comparable to or even surpassing the FSOL approach.
\end{itemize}

\section{Related works}
To address the issue of limited labeled data in new categories and dense scenarios, researchers have investigated various strategies. This section provides a concise overview of advancements in object localization, few-shot learning, and zero-shot learning.

\subsection{Object Localization}
With the spread of smartphones and cameras, a large number of images and videos have become interested in research. In the field of object localization, a variety of methods have emerged, including density map-based methods, improved resolution methods, and point-based methods \cite{20}. These methods have achieved certain results in different scenarios. For the tiny single instance problem, Ma et al. \cite{21} proposed a method to estimate the density map using a sliding window, where individual object positions are recovered directly from the estimated density map using integer programming while estimating the bounding box of each object. Wang et al. \cite{22} proposed the SCALNet counting and locating network, which achieves the localizations of each head in the crowd by means of the key-point locating network and by decoding the local peaks on the output heat map to obtain the location of the head. To solve the localization problem in crowded scenes, Liu et al. \cite{23} chose to improve the localizations accuracy by increasing the resolution through a cyclic attention scaling network, which recursively detects blurred image regions and converts them to high resolution for re-examination. In the medical domain, Tyagi et al. \cite{24} predicted location maps using a location network, which was then post-processed to obtain cell coordinates. In addition, Song et al. \cite{25} proposed a purely point-based framework for joint crowd counting and individual localizations by designing a point-to-point network (P2PNet) to directly predict the point proposals of the head in an image, achieving a one-to-one match between the proposal and the learning target, and achieving very superior localizations accuracy on the counting benchmark. 

Despite significant advances in object localization research, these methods lack timeliness and generalizability when faced with new localization tasks or new categories of interest. In particular, they rely heavily on a large amount of labeled data to train their models, which makes these methods ineffective in limited sample scenarios, especially when new categories need to be localized quickly.

\subsection{Few-shot Learning}
Few-Shot Learning (FSL) is a machine learning technique designed to train models using only a limited number of labeled samples, enabling effective learning and inference on new tasks or categories. The core challenge of few-shot learning is to achieve good generalization ability with a limited training dataset, which is widely used in tasks such as classification \cite{csvt4}, action recognition \cite{csvt3}, counting and localization.

Current few-shot counting and localization approaches focus on specific classes and objects. These include a two-stage training strategy \cite{26} and an uncertainty-aware few-shot object detector, which trains the latter to perform few-shot counting by generating pseudo ground truth bounding boxes from the former.  In addition to introducing datasets suitable for the task of few-shot counting \cite{27}, this effectively improves the efficiency of model training and the ability to generalize. Moreover, some researchers have also proposed similarity comparison modules and feature enhancement modules \cite{28}, which have made significant progress in the Few Shot Object Counting task by comparing the projected features of the support image and the query image, generating reliable similarity maps, and using these maps to enhance the features to make the boundaries between different objects clearer. The FSOL task\cite{fsol} has broadened the scope of research on few-shot learning. The authors of the task also identified intra-class variation of objects as a key issue affecting FSOL and an inevitable problem in the field of FSOL.

The innovative strategies mentioned above not only improve the performability of the model with limited labelled data, but also significantly increase its generalization capabilities. However, it is worth noting that few-shot learning itself relies to a certain extent on the number and quality of training samples, which is a limitation in real-world application scenarios as the actual samples collected are often difficult to meet its stringent requirements. This limitation has motivated researchers to further explore more extreme learning paradigms, such as zero-shot learning.

\subsection{Zero-shot Learning}
Zero-shot learning (ZSL) is an important technique in the field of deep learning to effectively learn and reason about unseen categories. Early research relied heavily on attribute learning for knowledge transfer between categories \cite{csvt2} . Lampert et al. \cite{Lampert} enabled models to transfer knowledge from known categories to unseen categories by describing objects as a set of features. Since then, research has gradually shifted to cross-modal approaches to address the zero-shot problem. Socher et al. \cite{Socher}  enable knowledge transfer to unseen categories by mapping images to an unsupervised semantic space where word vectors are implicitly combined with visual modalities.

In the field of zero-shot counting (ZSOC), recent advances have focused on embedding space approaches.  Kang et al. \cite{Kang} proposed an end-to-end zero-shot object counting framework that integrates semantically-conditioned cue fine-tuning, learnable affine transformations, and segmentation-aware jump connectivity to enhance the ability of model to perceive textual information. In the field of crowd counting, \cite{Wu} designed a domain-invariant and domain-specific crowd memory module, which can effectively separate the domain-invariant and domain-specific information in image features. In addition, \cite{Jiang} et al. implemented text-guided zero-shot object counting through a hierarchical patch-text interaction module to better guide the extraction of dense image features. Amini-Naieni et al. \cite{Amini-Naieni}  and Huang et al. \cite{Huang} utilized GroundingDINO \cite{dino} and SAM \cite{SAM}, respectively, as baseline models to improve the accuracy and generalization of the object counting task using multimodal inputs. These studies not only demonstrate the great potential of visual language models for zero-shot object counting, but also provide new insights for other downstream tasks such as object localization.

Although significant progress has been made in zero-shot object counting techniques, several key challenges remain. The first challenge is how to effectively integrate information from different modalities. Secondly, how to reduce the inter-category differences and enhance the ability of models to migrate and generalize across different domains constitutes another major challenge.

\section{Methodology}
In this section, we introduce the definition of ZSOL task and our ZSOL framework.
\subsection{Problem Definition}
In an object localization task, the goal is to accurately identify the specific location of a particular object in an image. The traditional input representation for an object localization task is usually an image, assuming the input image is \(I \in \mathbb{R}^{H \times W \times C}\), where \(H\), \(W\), and \(C\) represent the image’s height, width, and number of channels. The object localization task is to determine the location of a specific object based on the input image \(I\). The location of a specific object can be described using center point coordinates, key point coordinates, or a bounding box. Considering that our localization task will be applied to locate the specific position of an object, the output in this paper will utilize the center point.
% , while the key point format is more suitable for the task of predicting multiple feature points (e.g., joints of the human body or facial features) and the bounding box format is more suitable for the task of detecting an object that does not need to consider the exact position.
% The center point representation is as follows. 
Assume that the ground-truth coordinates of the object center in the image \(I\) are given by \(y = (x_c^*, y_c^*)\). The predicted coordinates from the model \(f(I; \theta)\) are denoted as \(\hat{y} = (x_c, y_c)\), where \(\theta\) represents the model parameters. The object localization task can then be formulated as:
\begin{equation}
\label{eq10}
\hat{y} = f(I; \theta) = (x_c, y_c).
\end{equation}

The previous section introduced that researchers are gradually recognizing the importance of improving sample utilization in sample-limited situations and have implemented a number of methods for solving the FSOL task, however, research on zero-shot localization task is still relatively scarce.
With the advent of transformer architectures, the emergence of large multimodal models such as CLIP has significantly advanced the progress of zero-shot learning tasks in computer vision. The model is able to understand and process information from different modalities by embedding images and texts into the joint feature space, thus performing effective object localization and recognition without specific samples, demonstrating amazing accuracy and powerful generalization ability in the field of zero-shot learning.

\begin{equation}
\label{eq1}
W_{i t}=f(I_{e}, T_{e}),
\end{equation}
As illustrated in Eq. (\ref{eq1}), the CLIP model initially inputs $n$ image-text pairs into the image encoder and text encoder, respectively, and extracts the visual embedding $\mathit{I_e}$ and the text embedding $\mathit{T_e}$. Following a brief processing stage, the $n$-dimensional joint multimodal embedding $\mathit{W_{it}}$ is generated by combining with the projection layer $\boldsymbol{\textit{f}}(\cdot)$, and the similarity scores between these embeddings are calculated. Finally, the computed results are subjected to cross-entropy processing in order to obtain loss values for the purposes of model optimization and training.

Inspired by the principles of multimodal approaches, we introduce an innovative task for zero-shot object localization. Different from the traditional definition of object localization, our task focuses on automatically identifying and localizing objects based on visual and linguistic information without explicit annotation. Specifically, the task can be defined as the process of localizing a specific object within an image, without access to relevant samples, through the utilization of prior knowledge of the model and the incorporation of prompt words. The task is mathematically defined as follows:
\begin{equation}
\label{eq13}
\hat{y}=f(I;\theta;P)=(x_c, y_c),
\end{equation}
where $\mathit{P}$ represents the prompt words that align with the image, this input feature is not available in traditional object localization methods.

\begin{figure}[!t]
\centering
\includegraphics[width=0.5\textwidth]{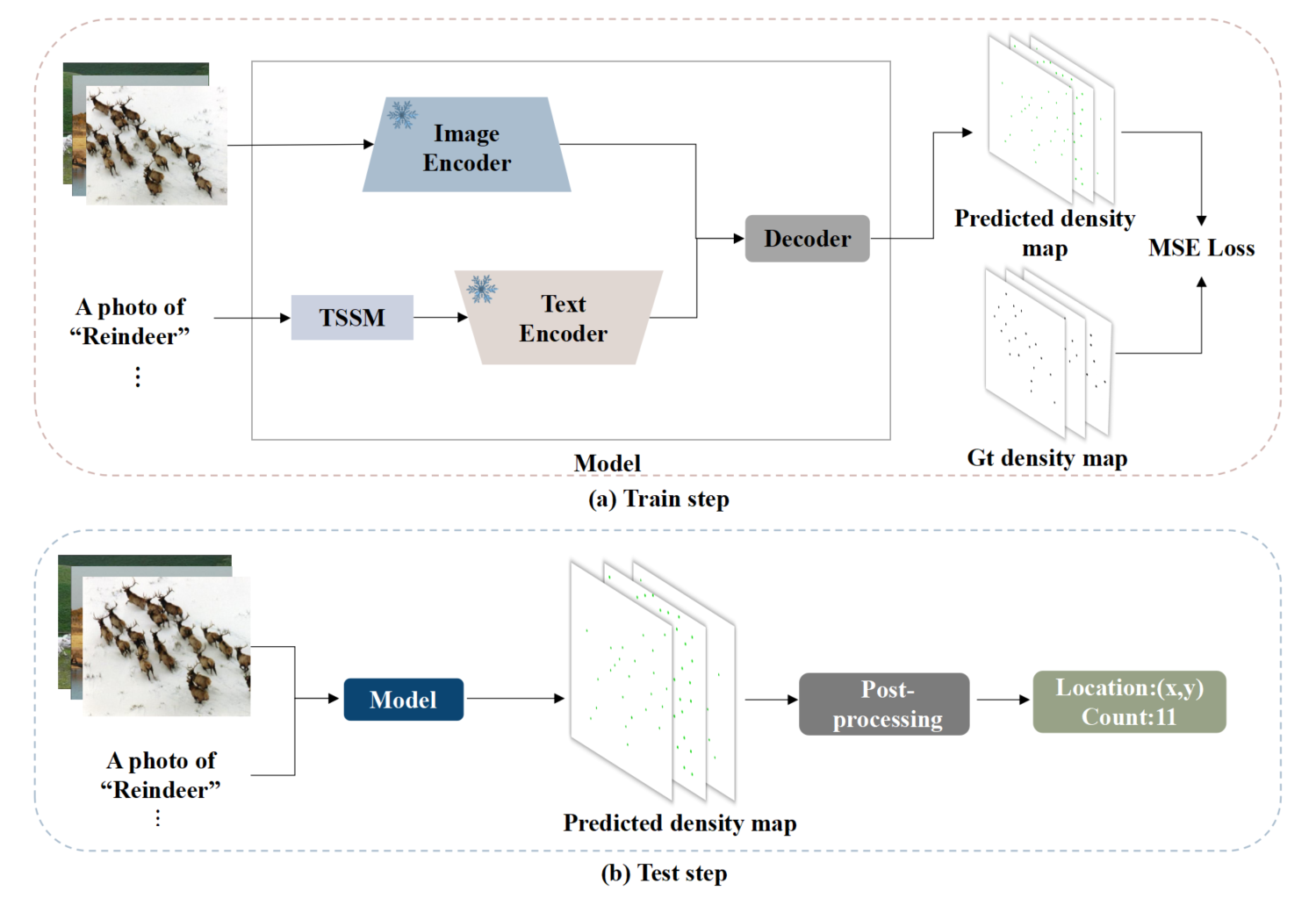}
\caption{Workflow for Zero-Shot Object Localization. (a) The encoder parameters are fixed, and the text self-similarity and visual features are extracted in conjunction with the TSSM module. (b) The density map generated by the model is subjected to post-processing. }
\label{fig_2}
\end{figure}

For the proposed ZSOL task, we introduce a solution concept, with the overall workflow illustrated in Fig. \ref{fig_2}. During the training stage, the parameters of the encoder are frozen. The visual encoder is responsible for extracting visual representations, while the TSSM module collaborates with the text encoder to extract the text representations. These two types of representations are subsequently mapped to a joint embedding space and transformed into a density map by the joint embedding decoder. The calculated Mean Squared Error (MSE) is used to optimize the model performance. In the testing stage, post-processing is applied to the output density map to extract the coordinates of the central points.

\subsection{ZSOLNet}
\begin{figure}[!t]
\centering
\includegraphics[width=0.5\textwidth]{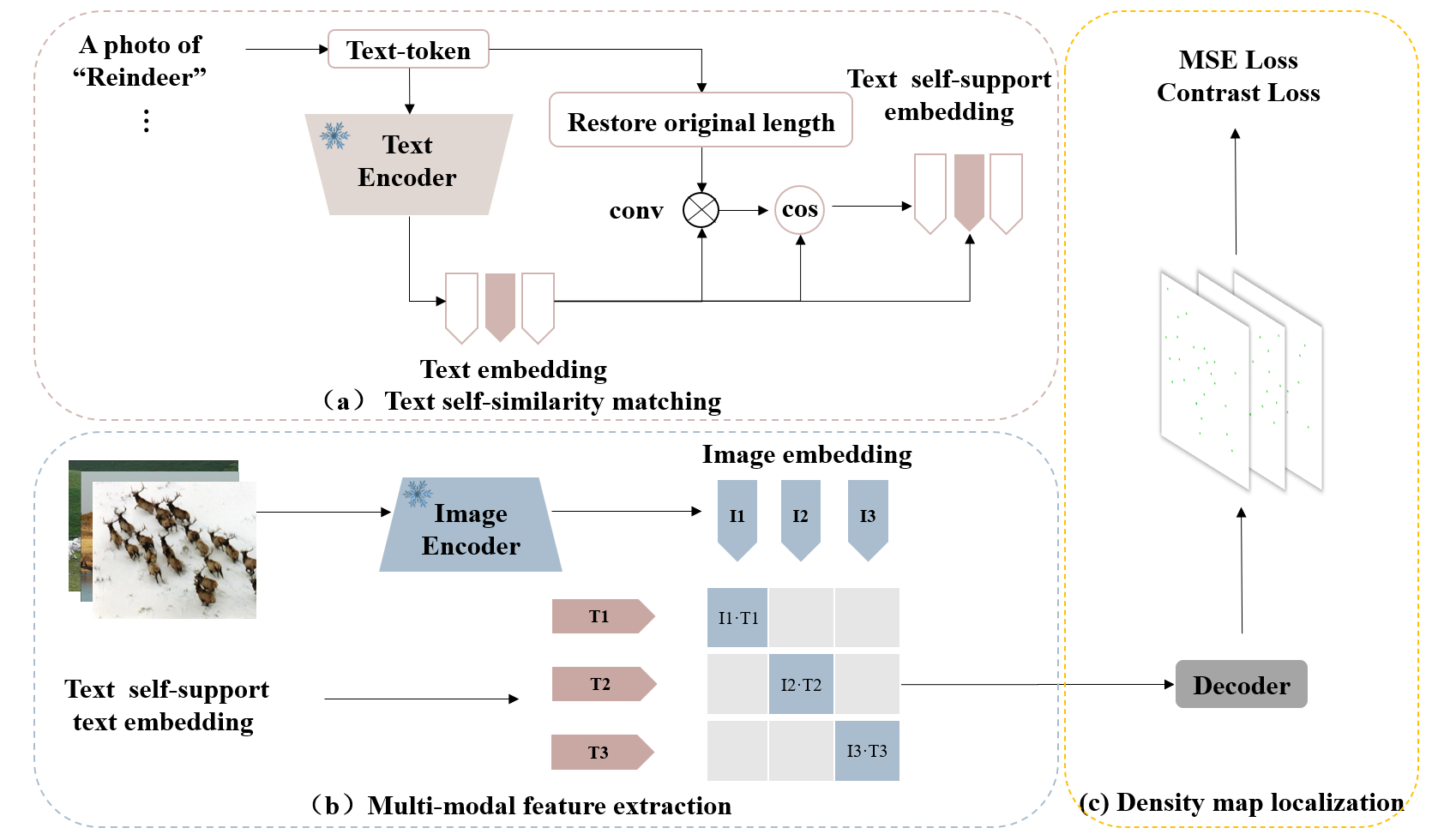}
\caption{The overall ZSOL framework. (a) Text self-similarity matching. Enhancing the weights of image-related features based on prompt word features. (b) CLIP-based feature extraction. Aligning text self-supporting embedding with image embedding. (c) Density map localization.}
\label{fig_3}
\end{figure}

The ZSOL model can be divided into three parts: text self-similarity matching, multi-modal feature extraction based on CLIP and density map localization. The framework of ZSOL model is shown in Fig. \ref{fig_3}.
% In this section, we will introduce the zero-shot object localization model in detail. The introduction can be divided into three parts: text self-similarity matching, multi-modal feature extraction based on CLIP and density map localization. The framework of ZSOL model is shown in Fig. \ref{fig_3}. 

\subsubsection{Text self-similarity matching}

% In order to achieve zero-shot object localization and reduce annotation costs, the pre-training knowledge of the CLIP model is used to freeze the majority of the parameters of the text encoder, with a small number of parameters being fine-tuned in order to maintain the basic representation ability of the model while enhancing adaptability. The prompt word is converted into a token and the embedding is obtained through the frozen text encoder. Subsequently, the correlation between the prompt word embedding and the original text embedding is calculated, and features that are closely related to the prompt word are extracted. The original embedding is then weighted based on these features to generate a self-supervised embedding of the text. Finally, this new embedding is used to fine-tune the model, effectively utilizing pre-training knowledge to reduce the dependence on annotation and improve the performance of zero-shot object localizations.
 
Although CLIP can achieve zero-shot object localization, it focuses more on extracting the overall information of the image. For example, there is an image of a man standing sideways in a corner of a large forest with a large black travel bag on his back. The task is to find and localize the man wearing a black T-shirt. The ZSOL model will misidentify the man whose upper body is mostly blocked by the black travel bag as the task object. Therefore, it is hoped that a module that can focus more on local details can be designed to improve the accuracy of localization.

\begin{figure*}[!t]
\centering
\includegraphics[width=0.7\textwidth]{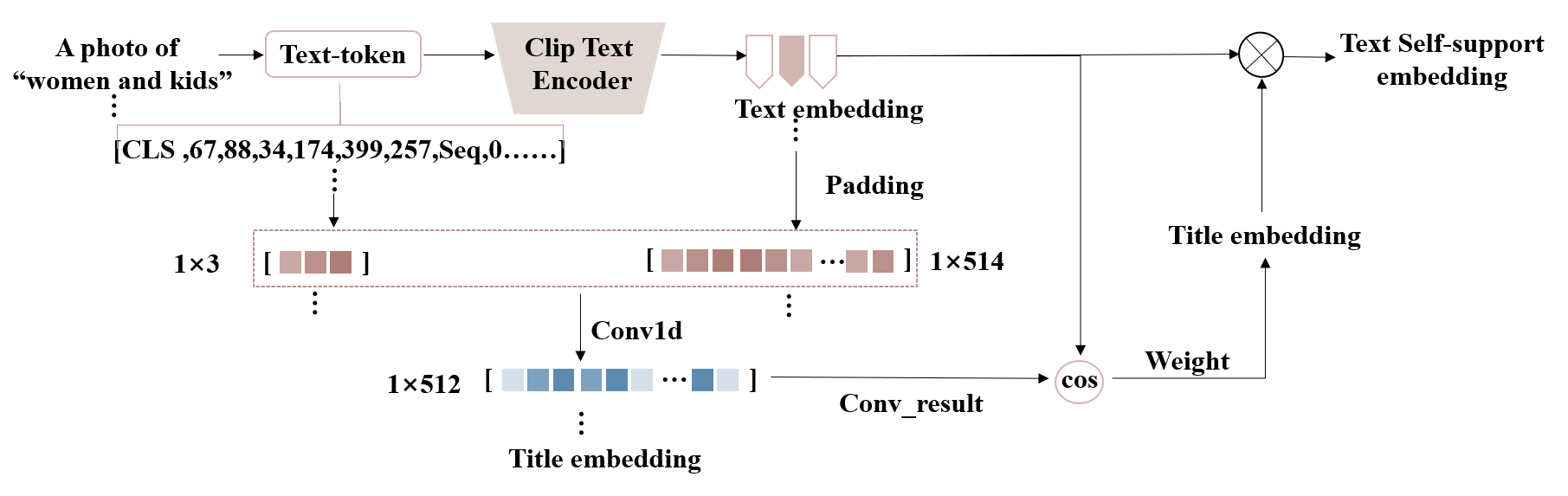}
\caption{Schematic diagram of TSSM module. Through text self-similarity matching and feature weighting, the recognition capability of the ZSOL model for local regions of interest is enhanced.}
\label{fig_5}
\end{figure*}

Inspired by Qi Fan et al. \cite{ssp}, we design a text self-similarity matching module based on the Gestalt theory of visual perception. The core of this module is that it can perform self-similarity matching between the features of the prompt word provided by the user and the features of the text after the text encoder, which not only better grasps the features of the object, but also improves the accuracy of object localization. Specifically, the prompt word is converted to token and the embedding is obtained by freezing the text encoder. Subsequently, the correlation between the prompt word embedding and the original text embedding is calculated, and the features that are closely related to the prompt word are extracted. The original embedding is then weighted according to these features to generate a self-supervised embedding of the text for fine-tuning the model. An example is shown in Fig. \ref{fig_5}.
% During the verification process, The title \( t \) of each image-text pair is expanded to ``A photo of \( t \)'', as shown in the figure.
The title ``women and kids'' is expanded to ``A photo of women and kids''. After the sentence has been segmented and mapped, the vector representation of the sentence in one-dimensional space is obtained. Add the special symbols [CLS] and [Seq] at the ends of this vector to indicate categories and sentence endings, and fill the rest of the positions with zeros to unify it into a vector of length 77. Obviously, in a sentence, the title is the main object. Therefore, we try to extract the mapping corresponding to the title from the text token as a kernel, which is used to further extract the features of the objects of interest from the text embedding. Since the title can be very long, we limit the length of the extracted mapping to no more than 3. The title embedding obtained by the convolution is again computed by cosine similarity with the text embedding, as shown in Eq. (\ref{eq2}).

\begin{equation}
\label{eq2}
W = \frac{\varepsilon_{t} \cdot \varepsilon_{o}}{\|\varepsilon_{t}\| \|\varepsilon_{o}\|} 
  = \frac{\sum_{i=1}^{n} \varepsilon_{t,i} \varepsilon_{o,i}}{\sqrt{\sum_{i=1}^{n} \varepsilon_{t,i}^2} \sqrt{\sum_{i=1}^{n} \varepsilon_{o,i}^2}},
\end{equation}
\begin{equation}
\label{eq9}
\varepsilon_{ts}=W*\varepsilon_{t} +\varepsilon_{o},
\end{equation}
where $\varepsilon_{t}$ represents the text features of the sentence after the text encoder, $\varepsilon_{o}$ represents the original text features of the prompt word, and W represents the cosine similarity between the two. \textit{n} = 3 represents three pairs of samples fed into each batch. The cosine similarity is used as a weight to weight the title embedding. The result is then weighted and summed with the text embedding to obtain $\varepsilon_{ts}$  $text$ $ self$ $support$ $ embedding$, as shown in Eq. (\ref{eq9}). This represents the feature representation that focuses on the local region of interest based on the overall features. Its size is equivalent to that of the text embedding, and it can be directly replaced and continue to participate in the verification step of the ZSOL model. The experimental data presented in Table 6 provides compelling evidence that this method significantly enhances the accuracy of zero-shot object localization.

\subsubsection{ Multi-modal feature extraction based on CLIP}

In the text self-similar matching module, the text self-similar matching embeddings are obtained, and in this subsection, the text self-similar embeddings and the corresponding images will be utilized as inputs for feature extraction. Similarly, based on freezing most of the parameters of the visual encoder, the input image is coded to generate the image embeddings. However directly using image embeddings for feature alignment may lead to the model overlooking local features. To address this issue, we introduce contrastive loss to guide the model in learning informative local visual representations, thereby enhancing its performance on dense prediction tasks. This approach uses a linear transformation to project the patch-level visual embedding into the same dimension as the text embedding, using contrast learning from the method to align image and text features, but it is worth noting that here positive patches are derived from the density map with ground truth, treating denser image patches as positive patches and less dense image patches as negative patches. With the aim of bringing the positive patch embedding and text embedding closer together, the model parameters are continuously updated until the patch embedding space is aligned with the text embedding space.

In delving into the process of extending the CLIP model from image-level classification tasks to dense prediction tasks such as density estimation, we adopt a strategy that is not limited to simply introducing contrast loss to enhance the ability of model to capture image details. To further refine this process, we design a multi-stage training mechanism aimed at balancing the learning of global features and local details of an image.

In the initial stage of training, we focus on shaping the model sensitivity to key patches (i.e., regions of higher density) in the image through contrast loss. At this stage, the model not only learns how to map each image patch to an embedding space that matches the text prompt, but also optimizes the richness and accuracy of the patch-level visual representation by maximizing the mutual information between positive patches and text embeddings, while minimizing the similarity between negative patches and text embeddings. This strategy drives the model to more accurately localize to regions in the image that are closely related to the density distribution of the object localization, laying a solid foundation for subsequent dense prediction tasks. At training stage, relying solely on contrast loss may cause the model to focus excessively on local details while ignoring the overall structure and contextual information of the image. Therefore, in the depth of training, we introduce MSE loss as a supplementary optimization objective. The calculation of MSE loss is based on the pixel-level difference between the model-predicted density maps and the real density maps, which helps the model to maintain the ability to recognize the global features of the image and ensure that the prediction results are both accurate and coherent. With this two-stage training strategy, we have achieved a good balance between image-level and patch-level feature learning by both exploiting the advantages of contrast loss for detailed capture and maintaining the model understanding of the overall structure of the image through MSE loss.

 % 在文本自相似匹配模块中，得到了文本自相似匹配嵌入，在这一小节里，将利用文本自相似嵌入和相应图像作为输入，基于clip的结构进行特征提取。类似的，在冻结 CLIP 模型中视觉编码器大部分参数的基础上，对输入图像进行编码，生成图像嵌入。然而，直接使用图像嵌入进行特征对齐，会使得模型忽略局部的特征，这个问题在上一小节已经举例说明了，因此还需要将图像级 CLIP 转移到密度估计等密集任务中。 我们引入了对比损失，引导模型学习信息丰富的补丁级视觉表示以进行密集预测。这种方法使用线性变换，将补丁级视觉嵌入投影到和文本嵌入相同的维度，使用对比学习从方法对齐图像和文本特征，但值得注意的是，这里的正补丁来自与接地真值密度图，将密度较高的图像补丁视作正补丁，密度较低的图像补丁视作负补丁。正补丁嵌入拉近固定文本嵌入的同时将负补丁推得更远，不断更新模型参数直到将补丁嵌入空间与文本嵌入空间对齐，具体公式可以参看clipcount。

%  在深入探讨将CLIP模型从图像级分类任务拓展至密度估计等密集预测任务的过程中，我们采取的策略不仅限于简单地引入对比损失以增强模型对图像细节的捕捉能力。为了进一步细化这一过程，我们设计了一套多阶段的训练机制，旨在平衡图像全局特征与局部细节的学习。

% 首先，在训练的初始阶段，我们专注于通过对比损失来塑造模型对图像中关键补丁（即密度较高区域）的敏感性。这一阶段，模型不仅学习如何将每个图像补丁映射到与文本提示相匹配的嵌入空间，还通过最大化正补丁与文本嵌入之间的互信息，同时最小化负补丁与文本嵌入的相似性，来优化补丁级视觉表示的丰富度和准确性。这种策略促使模型能够更精确地定位到图像中与目标密度分布紧密相关的区域，为后续的密集预测任务打下坚实基础。
% 随着训练的深入，仅仅依赖对比损失可能会导致模型过度聚焦于局部细节，而忽视了图像的整体结构和上下文信息。因此，在训练的深入，我们逐步引入均方误差（MSE）损失作为补充优化目标。MSE损失的计算基于模型预测的密度图与真实密度图之间的像素级差异，它有助于模型保持对图像全局特征的识别能力，确保预测结果既准确又连贯。通过这种双阶段训练策略，我们既利用了对比损失在细节捕捉方面的优势，又通过MSE损失维护了模型对图像整体结构的理解，实现了图像级与补丁级特征学习的良好平衡。

% 此外，为了进一步提升模型的泛化能力和稳定性，我们还在训练过程中引入了数据增强和正则化技术。通过对输入图像进行随机裁剪、旋转、缩放等操作，我们增加了训练样本的多样性，帮助模型学习到更加鲁棒的特征表示。同时，通过添加L2正则化项到损失函数中，我们有效抑制了模型参数的过拟合倾向，提高了模型的泛化性能。这些措施共同作用于训练过程，使得我们的模型能够在密度估计等密集预测任务中展现出更加优异的表现。

\subsubsection{ Density map localization and post-processing}
\begin{figure*}[!t]
\centering
\includegraphics[width=0.8\textwidth]{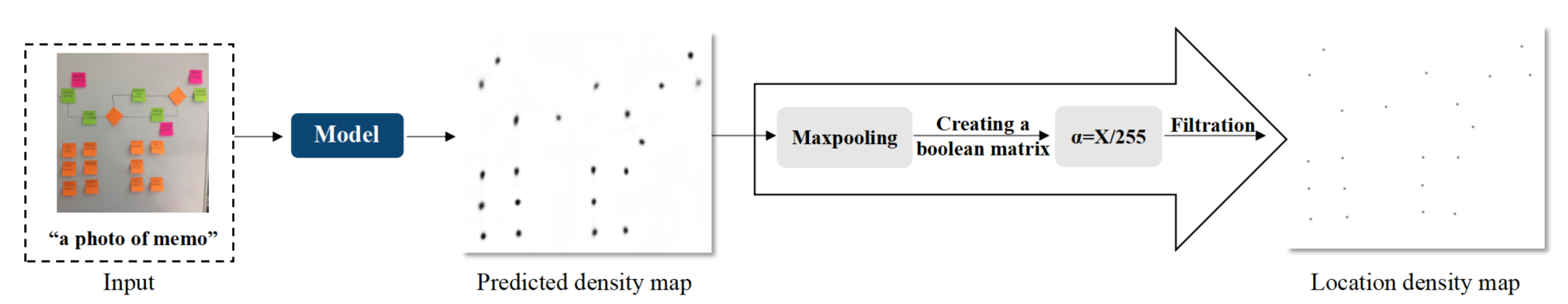}
\caption{Post-processing steps. By comprehensively applying pooling, filtering, and enhancement techniques to the coordinate points predicted by the density map model, the noise information in the image can be effectively weakened and eliminated. }
\label{fig_4}
\end{figure*}

 In the previous subsection, textual self-similarity matching embedding is implemented as a key step by extracting multimodal features after the ZSOL model and generating multimodal feature maps. In this subsection, we will introduce the post-processing module to decode and reduce the abstracted multimodal feature maps into density maps for generating the localization maps and obtaining the coordinates and number of objects in the natural images and calculating the contrast loss and MSE loss of the model and updating the model parameters. The main objective of post-processing is to detect local maxima in the image, which are considered as possible object locations. Drawing on the strategy of FIDTM \cite{FIDTM}, we introduce two adjustable thresholds $\alpha$ and $\beta$, where $\beta$ is the threshold used to determine the background noise, and $\alpha$ is the threshold used to identify peaks in the data, which can be adjusted in the experiments according to the sparseness and denseness of the data set. As shown in the Fig. \ref{fig_4}, the input density map is first passed through a maximum pooling operation of size 7x7 to aggregate and extract salient features in local regions. Subsequently, a new Boolean matrix is created in order to store and filter the aggregated feature information. Points with values greater than the preset threshold $\alpha$ are marked as positive samples, thereby reducing false positives and background noise, which in turn improves the accuracy of object detection.

Given that the zero-shot localizations task involves multiple datasets, it is evident that the object sizes in different datasets vary significantly. Therefore, the datasets are divided into two categories: dense and sparse, according to the average size of the objects in the dataset. For dense datasets, a lower threshold of $\alpha$ = 5/255 is set to capture the features of small objects more accurately, since the objects in these datasets are relatively small and closely distributed. In contrast, sparse datasets typically comprise larger, more sparsely distributed objects, necessitating an increased threshold of $\alpha$ = 10/255. This approach reduces the impact of differing object sizes while maintaining a certain degree of accuracy, thereby enhancing the adaptability of the zero-shot localization task to the characteristics of different datasets.
% 在上一小节中，文本自相似度匹配嵌入经过ZSOL模型提取出多模态特征并生成多模态特征图，实现了关键的一步。在这一小节，我们将介绍后处理模块将抽象的多模态特征图解码并还原成密度图，以便于生成定位图并获取自然图像中物体坐标和数量和计算模型的对比损失和MSE损失并更新模型参数。
% 后处理的主要目的是检测图像中的局部最大值，这些最大值被视为可能的目标位置。借鉴 FIDTM [35]的策略，我们引入了两个可调节的阈值α，和β，其中β是用来判定背景噪音的阈值，在实验中被设定为0.02，α是用来判定局部最大值点的阈值，在实验中可以根据数据集的稀疏和稠密程度进行调节。

Furthermore, in order to enhance the precision of positioning and minimize the impact of noise interference, a global lower threshold of $\beta$=0.06 is applied to the values of all candidate points. This implies that any candidate point with a value below 0.06 is regarded as a negative sample, indicating that the point is not part of the object position. 

% \subsection{Text Self-Similarity Matching}

% \begin{list}{}{}
% \item{\url{http://www.latex-community.org/}} 
% \item{\url{https://tex.stackexchange.com/} }
% \end{list}

\section{Experiments}
\subsection{Dataset}

To verify the effectiveness and generalization ability of text-guided zero-shot object localization, we selected three representative datasets for testing: FSC-147, CARPK and ShanghaiTech. These datasets not only cover multiple different categories of objects, but also provide rich annotation information, which provides strong support for our research on zero-shot object localization.

\subsubsection{FSC-147 dataset} It is a public dataset for few-shot counting tasks that has attracted much attention for its rich category information and fine annotations. The dataset contains 6,135 images of 147 different object categories, including 89 categories in the training set, 29 categories in the validation set, and 29 categories in the test set. The object instances are annotated with both point and bounding box annotations. This provides a valuable resource for exploring the localization of different category objects in zero-shot scenarios.

\subsubsection{CARPK dataset}It is a dataset of car park scenes with drone views, often used for counting tasks, with bounding box annotations. The dataset contains 1448 aerial images of car parks with a total of 89,777 annotated cars. Although the CARPK dataset focuses mainly on counting tasks, its fine car annotations make it suitable for zero-shot object localization research as well.

\subsubsection{ShanghaiTech dataset} It is a comprehensive crowd counting dataset with two parts: A and B. The images in Part A are randomly crawled from the web, most of which have large numbers of people. Part B is taken from the busy streets of Shanghai. It contains a total of 1,198 images with detailed annotations. The dataset is characterized by a diversity of viewpoints, providing an ideal platform for testing the generalization ability of the algorithms in zero-shot object localization tasks. By using the training data in Part A to localize the object population in Part B, we can evaluate the performance of the algorithm in different scenarios.

In summary, the FSC-147, CARPK and ShanghaiTech  datasets each have their own characteristics, providing us with a wealth of experimental materials and challenges for zero-shot object localization research.

\subsection{Implementation Details }
In this paper, the model is trained on the FSC-147 dataset based on the OpenAI pre-trained CLIP model and the ViT-B/16 backbone. During training, an epoch is defined as a batch of data fed into the model for training and validation. During the validation phase, the MSE loss and contrast loss were calculated to jointly evaluate the performance of the model. First, the model was pre-trained for 20 epochs using the contrast loss and then further trained for 200 epochs using the mean square error (MSE) loss. It is worth noting that in order to use training resources efficiently, only a portion of the image cropped from the full image was used during training, which may result in relatively low localization performance on the validation set. However, a sliding window approach was used in the testing phase, resulting in higher performance.

To train the model effectively, the AdamW optimizer was chosen as the main optimization algorithm. AdamW is a variant of the Adam optimizer that improves the generalization ability of the model by introducing a weight decay term to prevent overfitting. In addition, a learning rate scheduling strategy was used to further improve the training effect of the model. Specifically, the initial learning rate was set to 1e-4 and the learning rate was multiplied by 0.33 after every 100 training steps. This strategy allows the model to converge quickly in the early stages of training and to fine-tune the parameters by gradually reducing the learning rate in the later stages to achieve better performance. The entire training process took approximately 12 hours on a single Nvidia RTX-3090 Ti GPU.

\subsection{Evaluation Metrics}
In zero-shot object localization, since the number of points in the statistical density map can be used to count the number of objects, we use the performance evaluation metrics commonly used in localization and counting tasks to comprehensively measure the accuracy and effectiveness of the method. Next, we will introduce these evaluation metrics in detail to more scientifically evaluate the performance of the proposed method.

To measure the accuracy of object localization, we mainly use the following three key indicators

\begin{equation}
\label{eq3}
F 1=\frac{2\times\mathit{ Precision} \times \mathit{Recall}}{\mathit{Precision}+ \mathit{ Recall}},
\end{equation}
where $F1$ is the harmonic mean of precision and recall, which comprehensively considers the accuracy and coverage of the model.

\begin{equation}
\label{eq4}
AP = \sum_{r \in \{0, 0.01, \ldots, 1\}} P_{\text{interpolated}}(r) \times \Delta r,
\end{equation}
Where \( AP \) is defined as the sum of \( P_{\text{interpolated}}(r) \) multiplied by the interval \( \Delta r \), which measures the average precision of the model at different recall values. \( r \) represents the recall values interpolated in ascending order. \( AP \) indicates the model average precision across various recall rates, reflecting the proportion of correct predictions while maintaining a high recall rate. This metric directly assesses the recognition accuracy of the model for different types of objects.

\begin{equation}
\label{deqn_ex1a}
\mathit{AR}=\frac{1}{|T|} \sum_{t \in T} \mathit{ Recall}_{t},
\end{equation}
where $AR$  is the average recall rate of the model on different categories or objects. $AR$  is the average recall value, which is usually calculated by numerical integration or discretization. $T$ is the number of categories, and $\mathit{ Recall }_{t}$ is the recall value for category $t$.

\subsection{Result and Analysis}

Given the lack of direct comparison benchmarks, we have selected a series of current widely recognized methods as evaluation benchmarks to comprehensively measure the performance of our proposed ZSOL model. These methods cover fully-supervised localization methods, small-shot localization methods, and counting methods, specifically 10 fully-supervised localization methods: IIM\cite{38}, CLTR\cite{fsol}, TopoCount\cite{FIDTM}, LSC-CNN\cite{FIDTM}, TinyFace\cite{41}, RAZ\_Loc\cite{41}, FIDTM\cite{FIDTM}, GMS\cite{39}, STEERER\cite{41}, and CeDiRNet\cite{CeDiRNet}, and 7 few-shot methods: BMNet +\cite{40},CACL\cite{40}, Counting-DETR\cite{27}, LOCA\cite{LOCA}, FamNet\cite{ssrn}, Safecount\cite{FIDTM}, and FSOL\cite{fsol}. The parameter configurations and experimental conditions of all selected models can be traced back to their original published papers. It is worth noting that some of the methods, such as SafeCount and FamNet, were initially proposed as solutions for counting tasks, but were subsequently successfully applied to localization task by other researchers through post-processing means. Upon comparative analysis, these improved counting models outperform the original versions in both counting and localization performances, demonstrating good reliability. Therefore, these validated results are adopted in this paper as evaluation references and the specific years are marked in the table.

To further investigate the performance of zero-shot localization models on dense object localization tasks, experiments were conducted on three dense datasets, ShangHaiTech PartA, ShangHaiTech PartB and CARPK. In these experiments, the FIDTM and CLTR models used a low threshold $\sigma_s$ of 4 and a high threshold $\sigma _l$ of 8. To be consistent with these models, the ZSOL model also used the same threshold settings. For the GMS, CACL, BMNet+, IIM, and STEERER models, since the low threshold results were not provided in some papers, the high threshold setting used by these models, i.e., $\sigma_l = \sqrt{\frac{w^2 + h^2}{2}}$, where $w$ and $h$ represent the width and height of the image sample, respectively, was used.For models without a double threshold set, their SOTA results were used to compare with the high threshold performance of this paper.

\begin{table*}[!t]
\caption{Quantitative results of localization and counting indicators for the FSC-147 dataset\label{tab:table1}}
\centering
\resizebox{0.95\textwidth}{!}
{%
\tiny
\begin{tabular}{ccccccccccc}
\hline
\multirow{3}{*}{Supervised} & \multirow{3}{*}{Models} & \multirow{3}{*}{Backbone} & \multicolumn{8}{c}{Test set} \\ \cline{4-11} 
 &  &  & \multicolumn{3}{c}{Threshold=$\sigma_s(5)$} & \multicolumn{3}{c}{Threshold=$\sigma_l(10)$} & \multirow{2}{*}{MAE} & \multirow{2}{*}{MSE} \\ \cline{4-9}
 &  &  & F1 & AP & AR & F1 & AP & AR &  &  \\ \hline
\multirow{9}{*}{Few-Shot} & BMNet (2022)& ResNet-50 & - & - & - & - & - & - & 16.7 & 103.3 \\
 & BMNet+ (2022) & ResNet-50 & - & - & - & - & - & - & 14.6 & 91.8 \\
 & Counting-DETR (2024) & ResNet-50 & - & - & - & \underline{85.6} & \underline{85.2} & \underline{86.1} & 16.8 & 123.6 \\
 & LOCA (2023)& ResNet-50 & - & - & - & - & - & - &\underline{10.8} & \underline{57.0} \\
 & FamNet (2024)& ResNet-50 & - & - & - & 34.1 & 19.5 & 24.8 & 22.1 & 99.6 \\
 & Safecount (2024)& ResNet-18 & \underline{59.5} & - & - & 75.4 & - & - & 14.4 & 58.5 \\
 & FSOL (2024)& ResNet-50 & 48.9 & \underline{47.2} & \underline{50.7} & 70.7 & 68.3 & 73.4 & - & - \\
\hline
Zero-Shot & ZSOLNet & ViT-B & \textbf{51.1} & \textbf{48.7} & \textbf{53.6} & \textbf{65.1} & \textbf{62.1} & \textbf{68.4} & \textbf{18.7} & \textbf{106.6} \\ \hline
\end{tabular}%1
}
\end{table*}

% table1
% \begin{table}[]
% \centering
% \resizebox{\textwidth}{!}{%
% \begin{tabular}{cccccccccc}
% \hline
% \multirow{3}{*}{Supervised} & \multirow{3}{*}{Models} & \multicolumn{8}{c}{Test set} \\ \cline{3-10} 
%  &  & \multicolumn{3}{c}{Threshold=σs(5)} & \multicolumn{3}{c}{Threshold=σl(10)} & \multirow{2}{*}{MAE} & \multirow{2}{*}{MSE} \\ \cline{3-8}
%  &  & F1 & AP & AR & F1 & AP & AR &  &  \\ \hline
% \multirow{9}{*}{Few-Shot} & BMNet & - & - & - & - & - & - & 16.71 & 103.31 \\
%  & BMNet+ & - & - & - & - & - & - & 14.62 & 91.83 \\
%  & Counting-DETR &  &  &  & 85.6 & 85.2 & 86.1 & 16.79 & 123.56 \\
%  & LOCA & - & - & - & - & - & - & 10.79 & \textbf{56.97} \\
%  & FamNet & - & - & - & 34.1 & 19.5 & 24.8 & 22.09 & 99.55 \\
%  & Safecount & \textbf{59.48} & - & - & 75.42 & - & - & 14.36 & 58.46 \\
%  & FSOL & 48.9 & \textbf{47.2} & \textbf{50.7} & 70.7 & 68.3 & 73.4 & - & - \\
%  & DEN(ResNet-50) & - & - & - & 87.6 & 89.1 & 86.2 & 22.92 & 52.09 \\
%  & DEN(ViT-S) & - & - & - & \textbf{90.03} & \textbf{91.2} & \textbf{89.5} & \textbf{10.31} & 91.08 \\ \hline
% Zero-Shot & ZSOLNet & \textbf{45.0} & \textbf{50.6} & \textbf{40.5} & \textbf{55.2} & \textbf{62.12} & \textbf{49.75} & \textbf{20.78} & \textbf{100.14} \\ \hline
% \end{tabular}%
% }
% \end{table}

The quantitative results of the FSC-147 dataset positioning and counting indicators are shown in Table \ref{tab:table1}. The performance of the ZSOL model is compared with that of the most advanced FSOL model on the FSC-147 dataset. Although the quantitative results of the ZSOL model on the localization and counting metrics are slightly lower than those of the few-shot localization and counting model, it is worth emphasizing that the unique advantage of the zero-shot learning method is that it does not require additional labelled data. Nonetheless, the ZSOL model performance on the location and counting tasks is comparable to that of the few-shot location model, which fully demonstrates its effectiveness. Furthermore, the ZSOL model demonstrates high accuracy at both high and low thresholds, thereby corroborating the hypothesis that the method can maintain high accuracy when dealing with unseen categories.

\begin{table*}[!t]
\centering
\caption{Quantitative Results for the ShanghaiTechA Dataset Positioning and Counting Metrics\label{tab:table2}}
\resizebox{0.9\textwidth}{!}{%
\tiny
\begin{tabular}{cccccccccc}
\cline{1-9}
\multirow{3}{*}{Supervised} & \multirow{3}{*}{Models} & \multirow{3}{*}{Backbone} & \multicolumn{6}{c}{Test Set} \\ \cline{4-9}
 &  &  & \multicolumn{3}{c}{Threshold=\(\sigma_s(4)\)} & \multicolumn{3}{c}{Threshold=\(\sigma_l(8)\)} \\ \cline{4-9}
 &  &  & F1 & AP & AR & F1 & AP & AR \\ \cline{1-9}
\multirow{9}{*}{Full} & IIM (2023) & HRNet & - & - & - & 73.3 & 76.3 & 70.5 \\
 & CLTR (2022)& ResNet-50 & 43.2 & 43.6 & 42.7 & 74.2 & 74.9 & 73.5 \\
 & TopoCount (2023)& VGG-16 & 41.1 & 41.7 & 40.6 & 73.6 & 74.6 & 72.7 \\
 & LSC-CNN (2023)& VGG-16 & 32.6 & 33.4 & 31.9 & 62.4 & 63.9 & 61.0 \\
 & TinyFace (2023) & ResNet-101 & - & - & - & 57.3 & 43.1 & 85.5 \\
 & RAZ\_Loc (2023) & VGG-16 & - & - & - & 69.2 & 61.3 & 79.5 \\
 & FIDTM (2023)& HRNet & \underline{58.6} & \underline{59.1} & \underline{58.2} & \underline{77.6} & \underline{78.2} & \underline{77.0} \\
 & GMS (2023)& VGG-16 & - & - & - & 78.1 & \underline{81.7} & 74.9 \\
 & STEERER (2023) & HRNet & - & - & \textbf{-} & 79.8 & 80.0 & \underline{79.4} \\ \cline{1-9}
\multirow{1}{*}{Few-Shot} & FSOL(2024) & ResNet-50 & \underline{52.4} & \underline{58.4} & \underline{47.6} & \underline{69.6} & \underline{77.6} & \underline{63.1} \\ \cline{1-9}
Zero-Shot & ZSOLNet & ViT-B & \textbf{75.69} & \textbf{75.24} & \textbf{76.15} & \textbf{76.53} & \textbf{76.08} & \textbf{77.0} \\ \cline{1-9}
\end{tabular}%
}
\end{table*}

\begin{table*}[!t]
\caption{Quantitative results of localization and counting metrics for the ShanghaiTechB dataset\label{tab:table3}}
\centering
\resizebox{0.9\textwidth}{!}{%
\tiny
\begin{tabular}{cccccccccc}
\cline{1-9}
\multirow{3}{*}{Supervised} & \multirow{3}{*}{Models} & \multirow{3}{*}{Backbone} & \multicolumn{6}{c}{Test set} \\ \cline{4-9}
 &  &  & \multicolumn{3}{c}{Threshold=\(\sigma_s(4)\)} & \multicolumn{3}{c}{Threshold=\(\sigma_l(8)\)} \\ \cline{4-9}
 &  &  & F1 & AP & AR & F1 & AP & AR \\ \cline{1-9}
 \multirow{9}{*}{Full}& IIM (2023) & HRNet & - & - & - & 83.8 & 89.8 & 78.6 \\
 & FIDTM (2023)& HRNet & \underline{64.7} & \underline{64.9} & \underline{64.5} & 83.5 & 83.9 & 83.2 \\
 & TinyFace (2023) & ResNet-101 & - & - & - & 71.1 & 64.7 & 79.0 \\
 & LSC-CNN (2023)& VGG-16 & 29.5 & 29.7 & 29.2 & 57.0 & 57.5 & 56.7 \\
 & TopoCount (2023)& VGG-16 & 63.2 & 63.4 & 63.1 & 82.0 & 82.3 & 81.8 \\
 & RAZ\_Loc (2023) & VGG-16 & - & - & - & 68.0 & 60.0 & 78.3 \\
 & GMS (2023)& VGG-16 & - & - & - & 86.3 & \underline{91.9} & 81.2 \\
 & STEERER (2023) & HRNet & - & - & - & \underline{87.0} & 89.4 & \underline{84.8} \\ \cline{1-9}
\multirow{1}{*}{Few-Shot} 
 & FSOL (2024) & ResNet-50 & \underline{67.2} & \underline{75.5} & \underline{60.5} & \underline{78.0} & \underline{88.4} & \underline{70.9} \\ \cline{1-9}
Zero-Shot & ZSOLNet & ViT-B & \textbf{51.5} & \textbf{49.9} & \textbf{53.2} & \textbf{63.4} & \textbf{64.4} & \textbf{65.8} \\ \cline{1-9}
\end{tabular}%
}
\end{table*}

The quantitative results of the localization and counting metrics on the ShanghaiTechA and TechB datasets are presented in Table \ref{tab:table2} and Table \ref{tab:table3}. It is notable that, despite being based on zero-shot learning, the zero-shot localization model exhibits impressive localization performance on the ShanghaiTechA dataset, even surpassing some of the few-shot and fully supervised models on certain metrics. This outcome serves to illustrate the advantages of the zero-shot localization model in the context of dense crowd scenes. The ShanghaiTechA and ShanghaiTechB datasets are distinguished by the fact that the A dataset contains dense crowd scenes, while the B dataset reflects a sparse crowd distribution. A comparison of the experimental results on the two datasets reveals that the zero-shot localizations model performs more effectively in dense crowd scenes and is relatively less effective in sparse crowd scenes. This finding serves to reinforce the applicability of zero-shot object localizations in scenes characterized by dense objects.

\begin{table*}[!t]
\caption{Quantitative results for the CARPK dataset positioning and counting metrics\label{tab:table4}}
\centering
\resizebox{0.9\textwidth}{!}{%
\tiny
\begin{tabular}{ccccccccc}
\hline
 &  & \multirow{3}{*}{Backbone} & \multicolumn{6}{c}{Test set} \\ \cline{4-9} 
Supervised & Models &  & \multicolumn{3}{c}{Threshold=\(\sigma_s(5)\)} & \multicolumn{3}{c}{Threshold=\(\sigma_l(10)\)} \\ \cline{4-9} 
 &  &  & F1 & AP & AR & F1 & AP & AR \\ \hline
Full & CeDiRNet (2024)& ResNet-101 & - & - & - & \underline{96.2} & \underline{98.2} & \underline{94.7} \\ \hline
 & BMNet+ (2023)& ResNet-50 & - & - & - & 67.9 & 61.4 & 75.8 \\
 & CACL (2023) & ResNet-50 & - & - & - & 90.7 & 94.8 & 86.8 \\
 & FamNet (2024) & ResNet-50 & - & - & - & 43.8 & 41.2 & 46.7 \\ 
\multirow{1}{*}{Few-Shot} & DEN (2024) & ResNet-50 & - & - & - & 45.1 & 43.3 & 47.1 \\
 & DEN (2024)& ViT-S & - & - & - & 46.9 & 45.2 & 48.7 \\
 & FSOL (2024)& ResNet-50 & \underline{81.8} & \underline{80.9} & \underline{82.8} & \underline{93.5} & \underline{92.4} & \underline{94.6} \\ \hline
Zero-Shot & ZSOLNet & ViT-B & \textbf{87.7} & \textbf{91.6} & \textbf{84.2} & \textbf{88.6} & \textbf{92.5} & \textbf{85.1} \\ \hline
\end{tabular}%
}
\end{table*}

\begin{figure}[!t]
\centering
\includegraphics[width=0.5\textwidth]{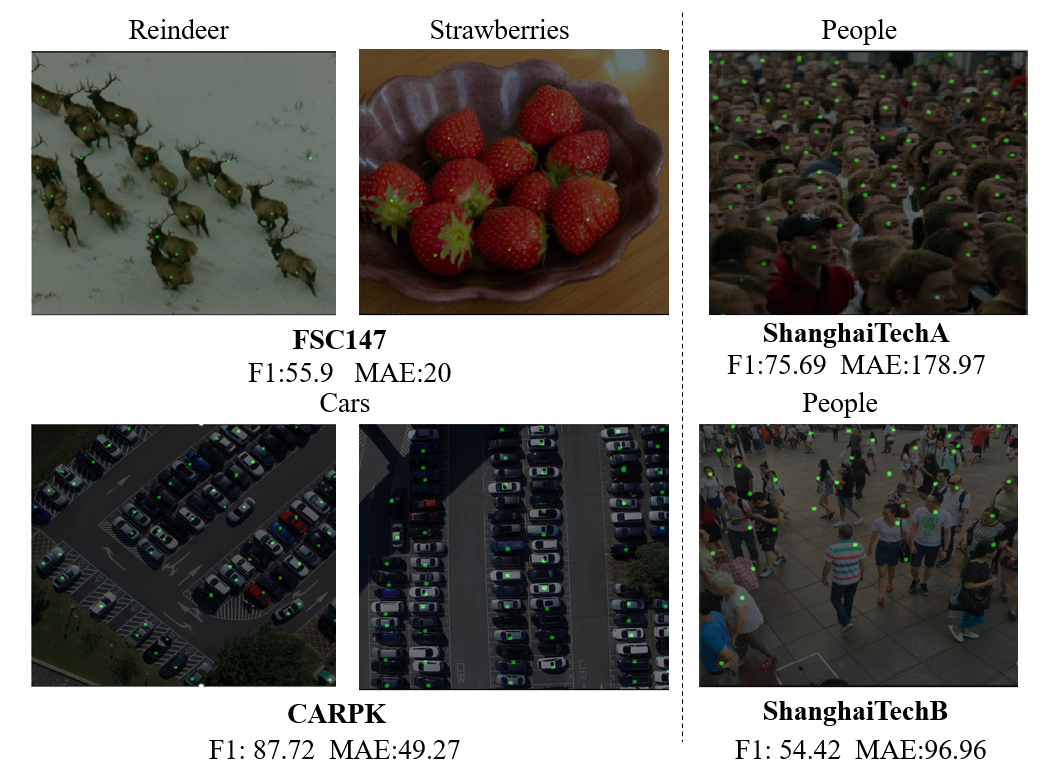}
\caption{The visualization results of the localization performance of the FSC-147 , ShanghaiA, ShanghaiB and CARPK datasets.}
\label{fig_6}
\vspace{-8pt}
\end{figure}

In order to ascertain the general applicability of the zero-shot localization model, further tests were conducted on the CARPK dataset. Table \ref{tab:table4} presents the quantitative results of the localization and counting metrics for the CARPK dataset.  The experimental results demonstrate that the zero-shot localization model maintains excellent localization performance at low thresholds, that is, when the evaluation criteria are relatively strict. This result not only proves the robustness of the model in dealing with challenging scenarios, but also further demonstrates the ability of the model to capture details. In addition, when the threshold is increased, that is, the evaluation criteria become relatively loose, the zero-shot localization model achieves a certain improvement in localization performance, even surpassing most of the few-shot models. The test results on the CARPK dataset fully verify the versatility of the zero-shot localization model in diverse scenarios. The zero-shot location model demonstrates excellent performance under both strict and loose evaluation criteria.
 Finally, to  demonstrate the performance of the ZSOL model, we visualized the qualitative and quantitative results of the ZSOL model on four datasets in Fig. \ref{fig_6}.

In particular, we overlay the density map predicted by the model with the original input image in order to visualize the prediction results. It should be noted that in order to improve the accuracy of the prediction and to cover different regions in the image, we used a sliding window approach for localized extraction of the image. Specifically, we set the size of the sliding window to 384x384 pixels and slid it in steps of 128 pixels. As a result, for each input image, we were able to obtain three predicted density maps at different locations, corresponding to the left, center, and right sides of the image. In this visualization, we only show the prediction results for the central region of each image to provide the most representative localization results. From the visualization results of the three datasets shown in Fig. \ref{fig_6}, we can observe that the ZSOL model exhibits excellent localization performance in scenarios where the object object localization is highly distinguished from the background. Meanwhile, when the  objects in the image are denser, the model is able to identify and localize these objects accurately. However, when faced with a background that has features similar to the color and texture of the object, or an image with a sparse distribution of objects, the localization performance of the ZSOL model is affected to a certain extent, and the likelihood of misclassification increases accordingly. These findings provide important directions for us to further improve and optimize the model.

\subsection{ Ablation Study}
In order to validate the effectiveness of the introduced modules and hyperparameter selection, ablation studies were performed on the validation and test sets of the FSC-147 dataset, respectively. Since the similarity map dimension of the TSSM module input is the same as the similarity map dimension of the output, it is only necessary to simply remove the TSSM module.

\begin{table*}[!t]
\caption{Evaluation of the validity of the TSSM in the FSC-147 dataset\label{tab:table5}}
\centering
\resizebox{0.6\textwidth}{!}{
\begin{tabular}{ccccccccc}
\hline
 & \multicolumn{8}{c}{Validation set} \\ \cline{2-9} 
TSSM & \multicolumn{3}{c}{Threshold=$\sigma_s(5)$} & \multicolumn{3}{c}{Threshold=$\sigma_l(10)$} &  &  \\ \cline{2-7}
 & F1 & AP & AR & F1 & AP & AR & MAE & MSE \\ \hline
$\sqrt{}$ & 23.93 & 22.34 & 25.77 & 34.81 & 32.50 & 37.48 & 28.46 & 133.60 \\
$\times$ & 20.22 & 18.33 & 22.54 & 31.59 & 28.63 & 35.22 & 24.33 & 83.26 \\ \hline
 & \multicolumn{8}{c}{Test set} \\ \cline{2-9} 
TSSM & \multicolumn{3}{c}{Threshold=$\sigma_s(5)$} & \multicolumn{3}{c}{Threshold=$\sigma_l(10)$} & &  \\ \cline{2-7}
 & F1 & AP & AR & F1 & AP & AR & MAE & MSE \\ \hline
$\sqrt{}$ & 45.00 & 50.60 & 40.50 & 55.20 & 62.12 & 49.75 & 20.78 & 100.14 \\
$\times$ & 40.42 & 47.17 & 35.63 & 48.80 & 50.65 & 46.01 & 18.21 & 105.43 \\ \hline
\end{tabular}
}
\end{table*}

Table \ref{tab:table5} illustrates the impact of the TSSM module on the model performance. The $\sqrt{}$ symbols indicate that the TSSM module has been incorporated into the model, whereas the × symbols indicate that the model represents a baseline without the TSSM module. The results of the TSSM ablation experiments demonstrate the significance of this module in the zero-shot object localization task. It is noteworthy that the performance of the validation set is inferior to that of the test set due to the fact that the images are cropped during training and validation, while the sliding window method is employed to process the images during retesting. Consequently, in the sub-ablation experiments, the data of the test set and validation set are compared separately. As illustrated in Table \ref{tab:table5}, the removal of the TSSM module results in a notable decline in localization performance, both in the validation set and the test set. Specifically, when the stricter threshold $\sigma=5$ is used for positive and negative sample determination, the F1-score of the validation and test sets decreases by 15.50 \% and 10.18\%, respectively. Additionally, the AP metric decreases by 17.95\% and 6.78\%, while the AR decreases by 12.53\% and 12.02\%, respectively. Moreover, when the threshold is relaxed to $\sigma=10$, although the determination of positive and negative samples becomes relatively relaxed, the performance loss caused by the missing TSSM module remains significant. In this case, the F1-score of the validation and test sets decreased by 9.25\% and 11.59\%, the AP decreased by 6.03\% and 7.52\%, while the decrease in AR was relatively small, 0.83\% and 1.89\%, respectively. These results fully demonstrate that the TSSM module has a positive effect on improving zero-shot object localization performance. Both under strict and loose threshold settings, TSSM can effectively improve the localization accuracy, confirming its effectiveness and importance in zero-shot object localization tasks.

\section{CONCLUSION}
 In this paper, we propose the ZSOL framework for the first time, which is able to achieve accurate object localization of objects only through prompt words without any labeled information, setting up a high-performance benchmark in the field of zero-shot localization. To enhance the semantic representation of prompt words, we design the TSSM module, which effectively improves the accuracy of the localization task. The ZSOL framework has demonstrated good performance on several standard datasets including FSC147, CARPK and ShanghaiTech, thus proving its suitability and effectiveness in handling complex tasks. Compared with traditional models, ZSOL demonstrates significant advantages in handling natural image object localization task with insufficient annotations.
 
 Although the text self-similarity matching module in the ZSOL framework has achieved significant results, we still recognise that there is room for improvement in its algorithmic optimization. In the future, we will work on the iterative upgrading of this module, and plan to introduce a horizontal enhancement network based on a two-branch structure and refine the self-attention mechanism to achieve more accurate recognition of object categories. This will further improve the compatibility of the ZSOL framework with various types of query images and broaden its application scenarios.

\newpage

\vspace{11pt}

\vspace{11pt}

\vfill

\end{document}